\documentclass{article}

\usepackage{PRIMEarxiv}

\usepackage[utf8]{inputenc} 
\usepackage[T1]{fontenc}    
\usepackage{hyperref}       
\hypersetup{
    colorlinks=true,
    linkcolor=blue,     
    citecolor=blue,     
    urlcolor=blue       
}
\usepackage{url}            
\usepackage{booktabs}       
\usepackage{amsfonts}       
\usepackage{nicefrac}       
\usepackage{microtype}      
\usepackage{lipsum}
\usepackage{fancyhdr}       
\usepackage{graphicx}       
\graphicspath{{media/}}     
\usepackage{hyperref}
\usepackage{array}
\usepackage{graphicx}
\usepackage{caption}
\usepackage{subcaption}
\usepackage{geometry}
\usepackage{amsmath}
\usepackage{amsmath}
\usepackage{amsfonts}
\usepackage{graphicx}

\title{\textbf{Exploring Advanced Large Language Models with LLMSuite}\\ \large \textit{A Free Comprehensive Recap of Techniques, Architectures, and Practical Applications}}
\author{
  Giorgio Roffo \\
  Sr. Research Scientist\\
  \texttt{\{giorgio.roffo\}@gmail.com} \\
}

\begin{document}
\maketitle

\begin{abstract}
This tutorial explores the advancements and challenges in the development of Large Language Models (LLMs) such as ChatGPT and Gemini. It addresses inherent limitations like temporal knowledge cutoffs, mathematical inaccuracies, and the generation of incorrect information, proposing solutions like Retrieval Augmented Generation (RAG), Program-Aided Language Models (PAL), and frameworks such as ReAct and LangChain. The integration of these techniques enhances LLM performance and reliability, especially in multi-step reasoning and complex task execution. The paper also covers fine-tuning strategies, including instruction fine-tuning, parameter-efficient methods like LoRA, and Reinforcement Learning from Human Feedback (RLHF) as well as Reinforced Self-Training (ReST). Additionally, it provides a comprehensive survey of transformer architectures and training techniques for LLMs. The source code can be accessed by contacting the author via email for a request.
\end{abstract}


\section{Introduction}
In recent years, the field of natural language processing (NLP) has witnessed groundbreaking advancements with the development and deployment of Large Language Models (LLMs) such as ChatGPT and Gemini. These models, characterized by their ability to generate human-like text, have set new benchmarks in a variety of applications, from automated customer support to creative writing. Despite their impressive capabilities, LLMs are not without limitations. They often struggle with the temporal limitations of their knowledge base, complex mathematical computations, and a tendency to produce plausible but incorrect information—commonly referred to as "hallucinations".

Addressing these limitations has become a focal point of contemporary research. One promising approach is the integration of LLMs with external data sources and applications, which can significantly enhance their accuracy and relevance without the need for costly retraining. Retrieval Augmented Generation (RAG) \cite{lewis2020retrieval} is a prominent technique in this regard, augmenting LLMs by linking them to up-to-date external databases to improve the precision of their outputs. Additionally, advanced prompting strategies like chain-of-thought prompting \cite{wei2022chain} aid in improving the reasoning capabilities of LLMs, especially for tasks requiring multi-step logic.

Moreover, specialized frameworks such as Program-Aided Language Models (PAL) \cite{gao2023pal} and ReAct \cite{yao2022react} have been developed to extend the functionalities of LLMs. PAL leverages external code interpreters to handle precise mathematical computations, while ReAct integrates reasoning with action planning to manage complex workflows. LangChain \cite{topsakal2023creating}, another significant development, offers modular components and agents to facilitate the integration of LLMs into diverse applications, enhancing their ability to execute complex tasks efficiently.

This tutorial paper\footnote{Special thanks to Shelbee Eigenbrode, Antje Barth, and Mike Chambers for their work on "Generative AI with Large Language Models" (Amazon AWS, 2023) \cite{coursera}, which significantly informed and inspired the material used in this document.}  provides a comprehensive exploration of the challenges and solutions associated with LLMs. It begins by examining the inherent limitations of LLMs, such as temporal knowledge cutoffs, mathematical inaccuracies, and hallucinations. The paper then introduces Retrieval Augmented Generation (RAG) as a means to access real-time external information, thus improving LLM performance across various applications. For instance, in customer service bots, RAG can enable real-time interactions with databases and APIs, enhancing the relevance and accuracy of responses.

The paper also discusses the integration of LLMs with external applications to perform complex tasks. Additionally, the concept of chain-of-thought prompting is introduced to enhance the reasoning capabilities of LLMs in multi-step tasks. This approach encourages the model to break down complex problems into intermediate steps, improving the coherence and logic of the generated responses.

Further, the paper delves into the Program-Aided Language Model (PAL) framework \cite{gao2023pal}, which pairs LLMs with external code interpreters to execute accurate calculations. This integration is crucial for improving the mathematical capabilities of LLMs, allowing them to handle tasks that require precise numerical computations. Recent advancements such as the ReAct \cite{yao2022react} framework and LangChain \cite{topsakal2023creating} are also explored, highlighting their potential in guiding LLMs through structured prompts to solve intricate problems and support advanced techniques like PAL.

The architectural components necessary for developing LLM-powered applications are thoroughly outlined, covering aspects such as infrastructure, deployment, and the integration of external information sources. The paper reviews various transformer-based models and recent advancements aimed at enhancing performance and efficiency. Techniques for scaling model training beyond a single GPU, including PyTorch’s Distributed Data Parallel (DDP) and Fully Sharded Data Parallel (FSDP) \cite{zhao2023pytorch}, are discussed, along with the ZeRO  \cite{rajbhandari2019zero} stages that optimize memory usage during training.

Fine-tuning strategies are a significant focus, exploring methods to enhance LLM performance for specific use cases. The paper examines instruction fine-tuning, multitask fine-tuning, and parameter-efficient fine-tuning (PEFT) methods like Low-Rank Adaptation (LoRA) \cite{hu2021lora} and prompt tuning \cite{lester2021prompt}. These techniques address the issue of catastrophic forgetting and suggest ways to mitigate it. Reinforcement Learning from Human Feedback (RLHF) \cite{ziegler2019,glaese2022improving} and Reinforced Self-Training \cite{gulcehre2023reinforced} are also explored as a method to align LLMs with human preferences. The use of Proximal Policy Optimization (PPO) \cite{schulman2017proximal} to update LLM weights based on human evaluations is discussed, along with challenges like reward hacking and the importance of maintaining model quality through techniques such as KL divergence.

Finally, the paper provides an introduction to PPO \cite{schulman2017proximal}, its phases, and its objectives, including policy loss, value function loss, and entropy loss. It discusses how PPO ensures stable learning by constraining updates within a trust region and its application in fine-tuning LLMs for human alignment. The integration of external data sources and the application of advanced frameworks and fine-tuning strategies significantly enhance the performance and reliability of LLMs, offering an overview of the current state and future directions in LLM-based applications.

The toolbox includes not only the code but also comprehensive tutorial slides to aid understanding and application. For fine-tuning models, we often need multi-task settings and diverse datasets to ensure robust performance across various tasks. Having references to the available datasets is crucial for researchers and practitioners in selecting appropriate data for their specific fine-tuning needs. Table \ref{tab:datasets} presents an overview of datasets and benchmarks commonly used in natural language processing tasks, including their type, citation, and a brief description. For additional resources and information, please contact or visit the \href{https://www.researchgate.net/profile/Giorgio-Roffo}{Author Page} or \href{https://scholar.google.it/citations?user=cD2LFuUAAAAJ&hl=en}{G. Roffo}. The source code can be accessed by contacting the author via email for a request.

In summary, this report serves as a guide to the techniques, architectures, and practical applications of Large Language Models. By addressing the inherent limitations of LLMs and introducing innovative frameworks and strategies, it aims to enhance the capabilities and reliability of these powerful tools in various real-world applications.

\section{Beyond Basic LLMs}

\begin{figure}[tbp]
    \centering
    \includegraphics[width=0.9\textwidth]{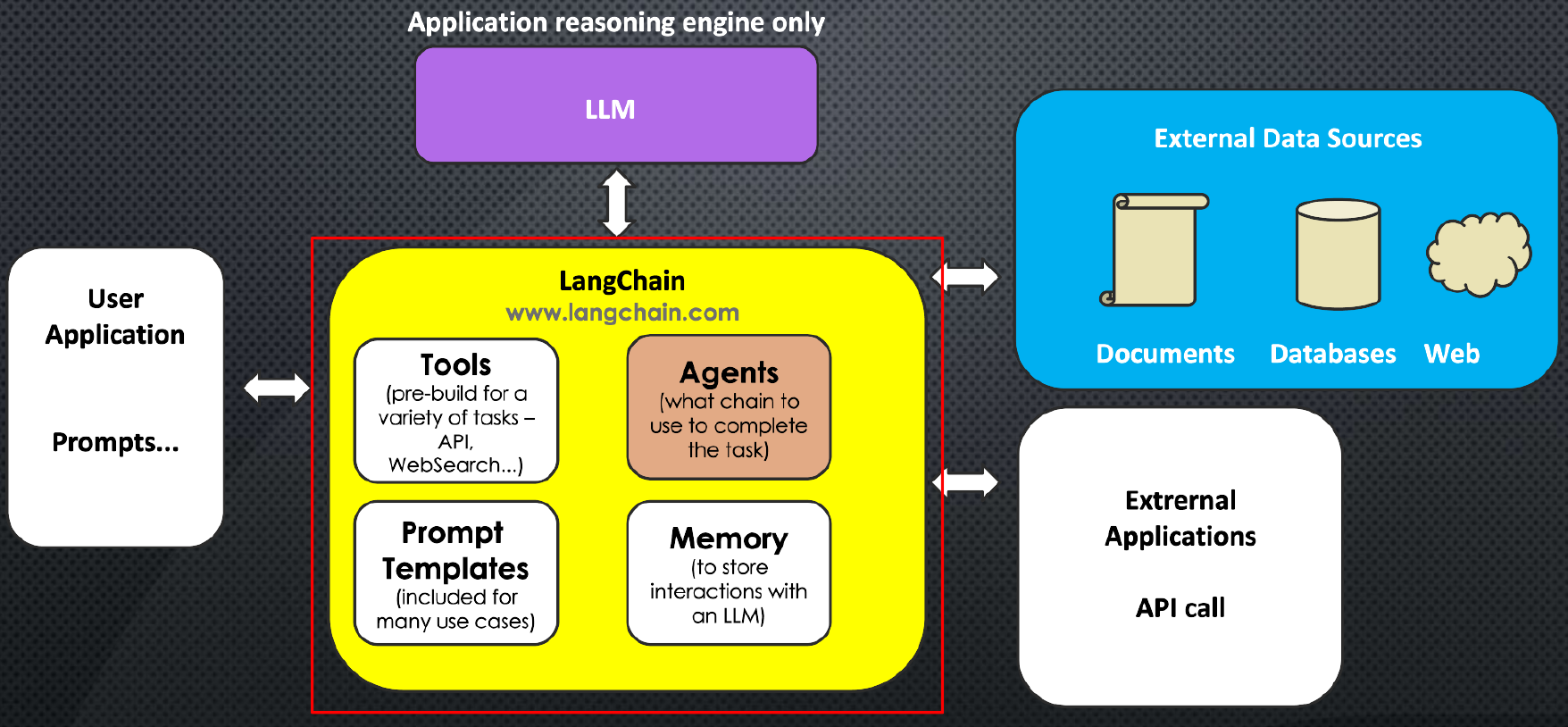}
    \caption{Overview of the framework including all components used to make an LLM application.}
    \label{fig:LLM_APP}
\end{figure}

The common perception that modern generative AI systems such as ChatGPT or Gemini are merely Large Language Models (LLMs) oversimplifies the sophisticated architecture underpinning these technologies. In reality, these systems integrate a multitude of frameworks and capabilities that extend far beyond the foundational LLMs. This section delves into the intricate ecosystem supporting these AI models, highlighting how they utilize external resources, advanced reasoning strategies, and dynamic problem-solving frameworks (see Fig.\ref{fig:LLM_APP}).

At the core of these advanced systems lies the LLM, which serves as the primary engine for generating human-like text. However, the LLM is just one component within a broader and more complex framework. For instance, tools like Retrieval-Augmented Generation (RAG) \cite{lewis2020retrieval} enhance the model's capabilities by enabling it to fetch information from external databases, wikis, or even real-time web searches. This integration allows the AI to provide more accurate and up-to-date responses by combining pre-trained knowledge with dynamically retrieved data.

Further enhancing the reasoning capabilities of these AI systems are techniques such as Chain of Thought (CoT) \cite{wei2022chain} and Program-Aided Language models (PAL) \cite{gao2023pal}. CoT enables the AI to break down complex queries into smaller, manageable steps, facilitating more coherent and logical responses. Similarly, PAL leverages external interpreters like Python to perform calculations or solve problems, which are then articulated in natural language by the LLM. This approach allows the AI to handle a wide range of tasks, from simple arithmetic to complex programming queries, by using the right tool for each specific task.

Another critical aspect of advanced AI systems is their ability to plan and execute strategies for problem-solving through frameworks like ReAct (Reasoning and Acting) \cite{yao2022react}. ReAct interleaves reasoning traces with task-specific actions, enabling the AI to generate and update action plans, handle exceptions, and interact with external sources for additional information. This enhances the model's adaptability, providing robust and flexible problem-solving capabilities. ReAct has proven effective in tasks like question answering and fact verification by reducing hallucinations and error propagation and has outperformed other methods in interactive decision-making benchmarks, ensuring improved interpretability and trustworthiness.

Frameworks such as GPT-4 All and LangChain \cite{pandya2023automating} (see Fig.\ref{fig:LLM_APP}) encapsulate these diverse functionalities into cohesive systems. These frameworks combine the powerful generative abilities of LLMs with external tools and advanced reasoning strategies, creating the illusion of a seamless, all-encompassing AI. By leveraging these complex systems, generative AI can deliver more sophisticated, accurate, and contextually relevant responses, moving closer to the ideal of a "perfect AI."

In summary, while LLMs are the foundation of generative AI, the true capabilities of these systems are realized through an intricate web of additional frameworks and tools. These components work in concert to provide a level of functionality and versatility that far exceeds the capabilities of standalone language models.

\subsection{Retrieval-Augmented Generation (RAG) Framework}

Despite advancements in training, tuning, and alignment techniques, large language models (LLMs) like ChatGPT and Gemini face inherent challenges that cannot be addressed by training alone. One significant issue is the temporal limitation of their knowledge base. This knowledge cutoff renders the model's information outdated. LLMs also struggle with complex mathematical tasks. When prompted to perform arithmetic operations, models may provide incorrect results as they attempt to predict the next token rather than execute precise calculations. Furthermore, LLMs are prone to hallucinations, generating plausible but incorrect information. 
\begin{figure}[tbp]
    \centering
    \includegraphics[width=0.9\textwidth]{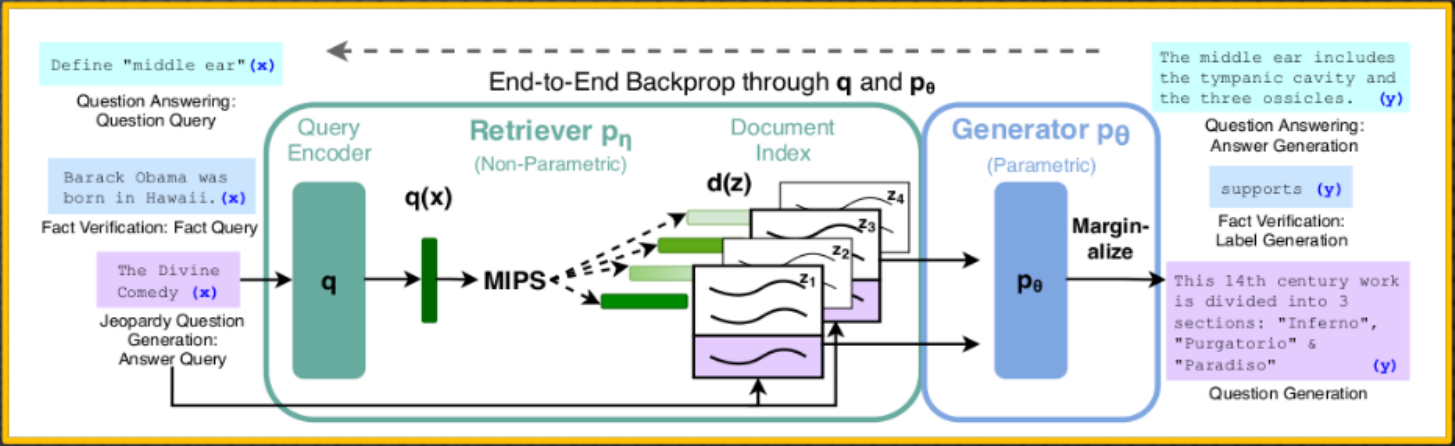}
    \caption{
        Retrieval-Augmented Generation (RAG) Framework. 
        \textbf{Components:} 
        1. \textit{Parametric Component (Generator):} A pre-trained seq2seq model (e.g., BART) generates responses using context from retrieved documents and the query.
        2. \textit{Non-Parametric Component (Retriever):} A dense vector index of documents (e.g., Wikipedia) acts as retrievable memory, with a neural retriever (e.g., DPR) fetching relevant documents based on the query.
        \textbf{Workflow:}
        1. \textit{Query Input:} The retriever processes the input query to find relevant context.
        2. \textit{Document Retrieval:} The retriever computes vector representations of the query and documents, retrieving the most relevant ones using techniques like Maximum Inner Product Search (MIPS).
        3. \textit{Sequence Generation:} The retrieved documents, along with the original query, are fed into the seq2seq generator, which produces the output text by integrating information from both sources.
    }
    \label{fig:RAG}
\end{figure}
To overcome these limitations, integrating external data sources and applications at inference time proves essential. Retrieval Augmented Generation (RAG) \cite{lewis2020retrieval} is a framework designed to connect LLMs to external data, thus updating their knowledge base and improving accuracy without costly retraining. By accessing current information from external databases, vector stores, or APIs, models can generate more relevant and precise responses. RAG involves a retriever component that encodes user queries and searches an external data source for relevant information. This augmented query is then processed by the LLM to produce a more accurate completion. For example, in legal applications, RAG can enhance discovery phases by querying a corpus of legal documents, thus retrieving pertinent information efficiently. Figure \ref{fig:LLM_APP} illustrates the architecture of an LLM-powered application using LangChain, demonstrating how various components such as tools, agents, and memory modules are integrated to facilitate complex task execution. 

This framework not only updates the model's knowledge but also mitigates hallucinations by grounding responses in external data. Implementing RAG requires careful handling of context window limitations and efficient data retrieval using techniques like vector stores, which facilitate rapid semantic searches.  In Figure \ref{fig:RAG}, we illustrate the Retrieval-Augmented Generation (RAG) framework, detailing its components and operational workflow to enhance the accuracy and reliability of generative AI models.

In conclusion, while LLMs face challenges like knowledge cutoffs, mathematical errors, and hallucinations, integrating external data sources through frameworks like RAG \cite{lewis2020retrieval} can significantly enhance their performance, making them more reliable and useful in various applications.

\subsection{Interactions of LLMs with External Applications}

Large Language Models (LLMs) like ChatGPT and Gemini are increasingly integrated with external applications to perform complex tasks. Connecting LLMs to external applications enables them to interact with the broader world, performing actions based on user inputs. This interaction is driven by prompts and completions, where the LLM serves as the reasoning engine. For effective operation, the completions must contain actionable instructions, formatted to be understood by the application. 

Despite their capabilities, LLMs often face challenges with complex reasoning, particularly in multi-step tasks or mathematical problems. An example is an LLM tasked with solving a simple multi-step math problem, where it incorrectly calculates the result due to inadequate reasoning steps. Researchers address this by employing "chain of thought" prompting, which encourages the model to break down problems into intermediate steps, mimicking human problem-solving processes \cite{wei2022chain}. This method has shown success in improving the model’s reasoning and accuracy in tasks such as arithmetic and physics problems. However, LLMs’ limited math skills can still hinder tasks requiring precise calculations. To overcome this, LLMs can be integrated with specialized programs that excel in mathematical computations. This approach ensures accurate outcomes for tasks like calculating sales totals, taxes, or discounts on e-commerce platforms. However, while chain-of-thought prompting can enhance the reasoning process, it does not guarantee accuracy in individual calculations, particularly with larger numbers or intricate operations \cite{wei2022chain}. This limitation is critical in applications where precise calculations are necessary, such as financial transactions or recipe measurements.

To address these limitations, LLMs can be augmented with external tools adept at handling mathematical computations. One prominent framework in this context is the Program-Aided Language Model (PAL), introduced by Luyu Gao and collaborators at Carnegie Mellon University in 2022 \cite{gao2023pal}. PAL pairs an LLM with an external code interpreter, typically a Python interpreter, to perform calculations. The PAL framework employs chain-of-thought prompting to generate executable Python scripts. These scripts, created by the LLM based on reasoning steps, are then executed by the interpreter. This method ensures accurate computations by leveraging the interpreter's mathematical capabilities.

\begin{figure}[tbp]
    \centering
    \includegraphics[width=0.9\textwidth]{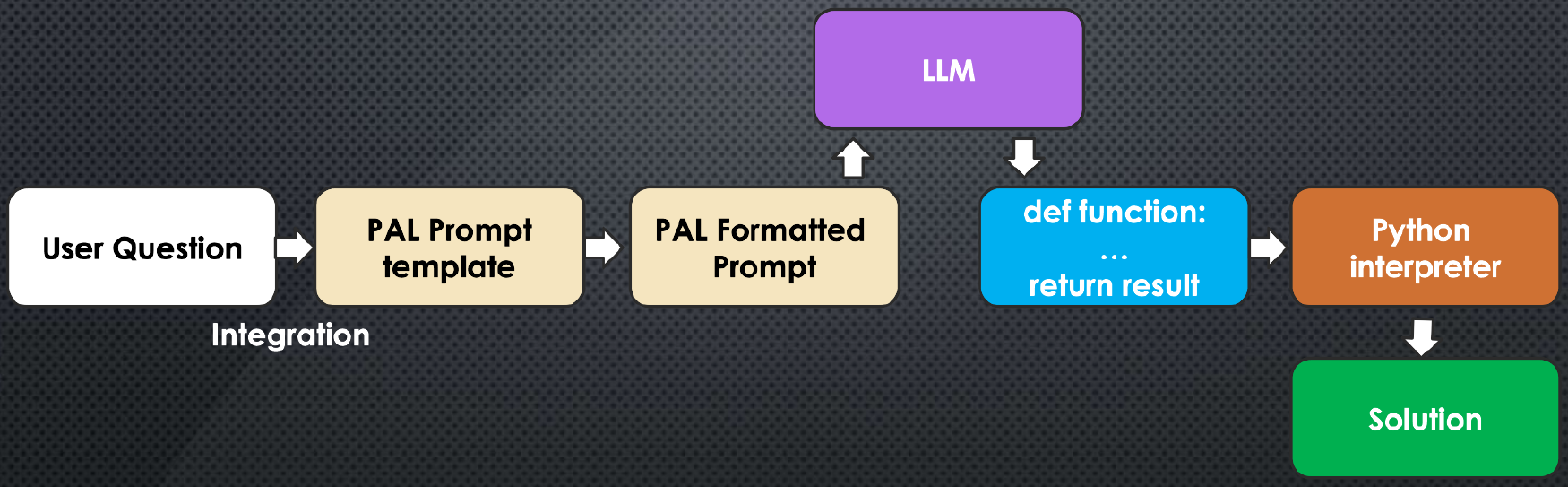}
    \caption{Pipeline of the Program-Aided Language Model (PAL) demonstrating the integration of user questions through PAL prompt templates and Python interpreters.}
    \label{fig:PAL}
\end{figure}

An illustrative example involves the model reasoning through a problem and generating corresponding Python code. Variables are declared and assigned values derived from the reasoning steps. The Python interpreter executes this code, ensuring the accuracy of the calculations. This process is demonstrated with an example where the LLM determines the remaining loaves of bread in a bakery after accounting for sales and returns. As shown in Figure \ref{fig:PAL}, the PAL pipeline integrates user questions through PAL prompt templates and Python interpreters, illustrating how this framework enhances the accuracy of LLM computations.

To implement PAL, a prompt is formatted to include one or more example questions followed by reasoning steps and Python code that solve these problems. The new question to be answered is appended to this prompt. When passed to the LLM, the model generates a completion in the form of a Python script. This script is then executed by a Python interpreter, and the resulting accurate answer is appended to the prompt. The updated prompt, containing the correct answer, is fed back into the LLM to generate a completion that includes the correct answer. For more complex mathematical problems, PAL ensures the accuracy and reliability of calculations. This framework can be automated using an orchestrator, which manages the flow of information between the LLM and external tools. The orchestrator, functioning as a control unit, directs the execution of actions based on the LLM's output, enhancing the application's overall reliability and efficiency.

In summary, while LLMs have inherent limitations in performing mathematical operations, augmenting them with frameworks like PAL and external interpreters significantly enhances their capability to handle complex computations accurately.

\subsection{ReAct Framework for Complex Problem Solving}

ReAct, developed by researchers from Princeton University and Google, stands for Reasoning + Acting. This framework enhances large language models (LLMs) by combining reasoning with actionable outputs, enabling them to interact with external tools and environments \cite{yao2022react}. ReAct allows LLMs to perform human-like operations by generating reasoning traces and taking actions to gather information, thus addressing limitations of Chain-of-Thought (CoT) prompting, such as fact hallucination, by integrating reasoning with real-world interactions. 

The ReAct framework is particularly effective in knowledge-intensive tasks like multi-hop question answering and decision-making in simulated environments. It demonstrates how structured examples can guide an LLM to reason through problems and decide on actions that advance toward a solution. ReAct's prompting strategy involves a thought-action-observation cycle. For instance, an example prompt might start with a question requiring multiple steps to answer, such as determining which of two magazines was created first. The "thought" represents a reasoning step, guiding the model to identify an action. The "action" uses a specific format to interact with external applications or data sources, such as searching Wikipedia. The "observation" integrates new information from the action back into the prompt, allowing the model to iterate until the final answer is determined. ReAct has proven especially useful in applications requiring external tool integration, exemplified by Microsoft's Office 365 Copilot. By allowing LLMs to generate verbal reasoning traces and actions for tasks, ReAct enhances their decision-making processes. Actions in the ReAct framework are limited to a predefined set, such as searching for Wikipedia entries, looking up specific keywords, or concluding the task. This restriction ensures that LLMs operate within the application's capabilities, preventing the proposal of unfeasible steps. Detailed instructions are provided to the model, defining the task and the allowable actions, ensuring robustness and reliability in operation.

In recent developments, structured prompts have shown significant potential in guiding LLMs to write Python scripts for solving complex mathematical problems. The Program-Aided Language Model (PAL) links LLMs to a Python interpreter, enabling the execution of code and returning results to the LLM \cite{gao2023pal}. However, most applications demand LLMs to manage more complex workflows involving multiple external data sources and applications. The ReAct framework addresses this need by combining chain-of-thought reasoning with action planning, exemplified through problems from HotPot-QA and FEVER benchmarks, which require multi-step reasoning over multiple Wikipedia passages.

\subsection{LangChain for Building LLM Applications}

LangChain is an open-source framework designed to streamline the development of applications based on Large Language Models (LLMs) \cite{topsakal2023creating}. It provides tools and abstractions to enhance the customization, accuracy, and relevancy of the information generated by LLMs. LangChain includes modular components such as prompt templates, memory storage, and tools for various tasks, such as API calls and external dataset access. These components enable developers to create optimized prompt chains and customize existing templates.

LangChain is particularly important for repurposing LLMs for domain-specific applications without retraining or fine-tuning. It allows development teams to build complex applications that reference proprietary information to augment model responses. For example, LangChain can be used to create Retrieval Augmented Generation (RAG) workflows that introduce new information to the language model during prompting, reducing model hallucination and improving response accuracy.

By abstracting the complexity of data source integrations and prompt refining, LangChain simplifies AI development, allowing developers to build complex applications quickly. Instead of programming business logic, software teams can modify templates and libraries provided by LangChain, reducing development time. LangChain supports agents that provide flexibility in deciding actions based on user input and can incorporate frameworks like PAL \cite{gao2023pal} and ReAct \cite{yao2022react}.

\section{Survey of Transformer Architectures in Language Models}

The transformer architecture has revolutionized natural language processing (NLP) by dramatically improving the performance of language models compared to earlier recurrent neural networks (RNNs). This improvement is largely due to the transformer's self-attention mechanism, which allows the model to consider the relevance of each word in a sentence to every other word, rather than just to its immediate neighbors. This capability enables the model to capture complex dependencies and contextual information effectively \cite{vaswani2017attention}. The transformer model comprises two main components: the encoder and the decoder, both of which consist of multiple layers. Each layer includes a self-attention mechanism followed by a feed-forward neural network.
\begin{figure}[tbp]
    \centering
    \includegraphics[width=0.5\textwidth]{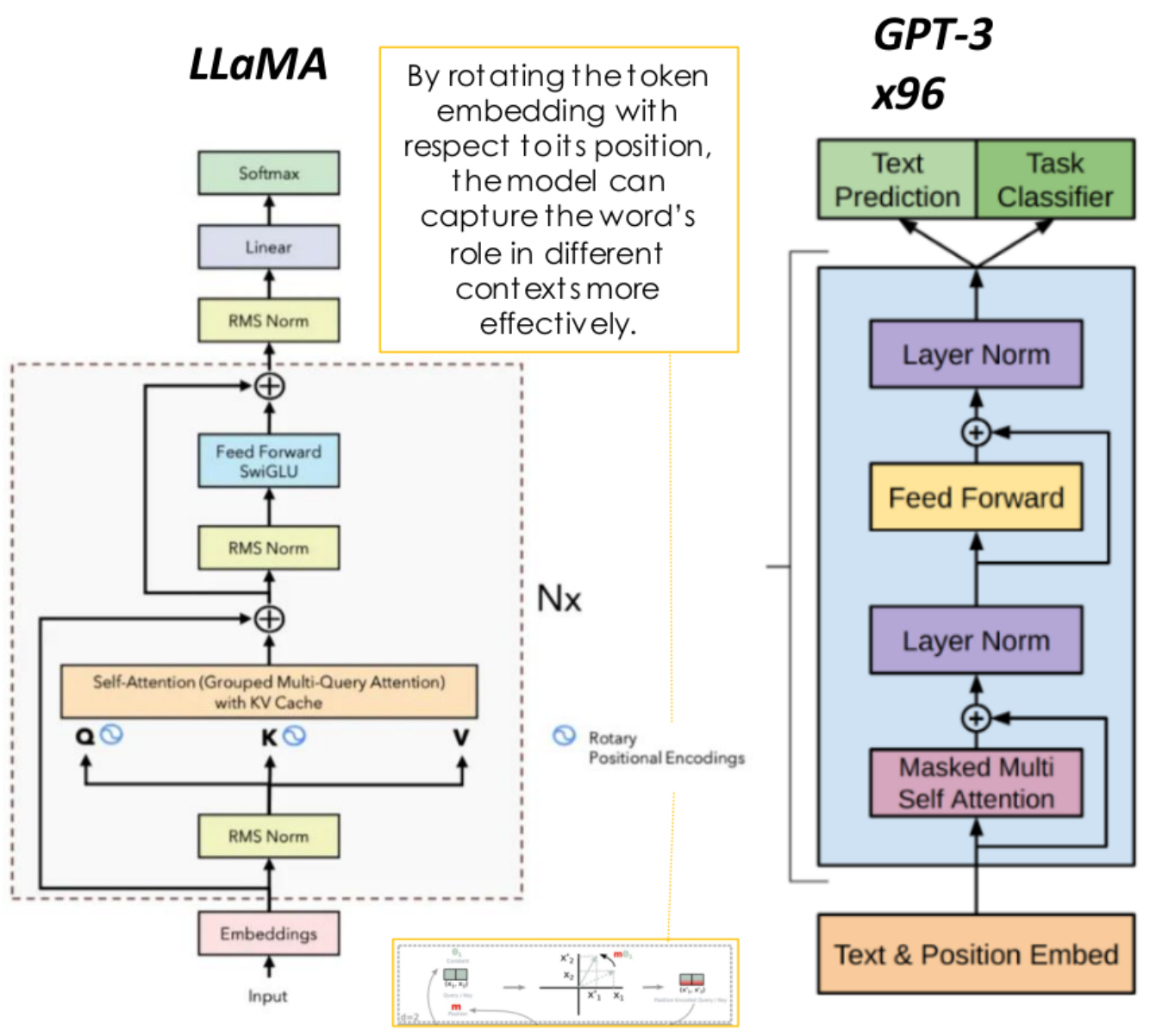}
    \caption{Comparison of LLAMA and GPT-3 (decoder-only) Architectures. The diagram on the left illustrates the LLAMA architecture, which incorporates a series of components including embeddings, rotary positional encodings, self-attention mechanisms with key-value caching, and feed-forward layers with RMS normalization. Notably, the LLAMA architecture utilizes grouped multi-query attention for efficient processing. On the right, the GPT-3 architecture is shown with its 96-layer deep structure featuring masked multi-self-attention, layer normalization, and feed-forward layers. The text and position embeddings are essential for initial input processing. A key insight highlighted is the use of token embedding rotation in LLAMA to effectively capture contextual word roles.}
    \label{fig:architectures}
\end{figure}
Before any processing, input text is tokenized into discrete tokens, which are then converted into numerical representations called token IDs. These token IDs are embedded into high-dimensional vectors in an embedding layer. In the original transformer architecture, the embedding dimension is typically 512 \cite{vaswani2017attention}. The self-attention mechanism calculates attention scores for each pair of tokens in the input sequence. These scores determine how much focus each token should give to every other token. Mathematically, given an input sequence of tokens represented as a matrix $X$ of shape $(N, d_{model})$, where $N$ is the sequence length and $d_{model}$ is the embedding dimension, the self-attention is computed using:
\[
\text{Attention}(Q, K, V) = \text{softmax}\left(\frac{QK^T}{\sqrt{d_k}}\right)V
\]
where $Q$, $K$, and $V$ are the queries, keys, and values matrices, all derived from $X$ through learned linear transformations \cite{vaswani2017attention}. Transformers utilize multi-headed self-attention, where the input is linearly projected into $h$ different subspaces (heads). Each head independently applies the self-attention mechanism, and their outputs are concatenated and linearly transformed. This allows the model to capture various aspects of the relationships between tokens:
\[
\text{MultiHead}(Q, K, V) = \text{Concat}(\text{head}_1, \ldots, \text{head}_h)W^O
\]
where each head $i$ is computed as $\text{head}_i = \text{Attention}(QW_i^Q, KW_i^K,
VW_i^V)$, with $W_i^Q$, $W_i^K$, and $W_i^V$ being the learned projection matrices for the $i$-th head \cite{vaswani2017attention}. Since transformers process all tokens in parallel, positional encoding is added to the input embeddings to retain information about the order of the tokens in the sequence. The positional encodings are added to the embeddings before they are fed into the self-attention layers:
\[
PE_{(pos, 2i)} = \sin\left(\frac{pos}{10000^{2i/d_{model}}}\right)
\]
\[
PE_{(pos, 2i+1)} = \cos\left(\frac{pos}{10000^{2i/d_{model}}}\right)
\]
where $pos$ is the position and $i$ is the dimension \cite{vaswani2017attention}.

Transformer models are trained on large datasets to learn the attention weights, enabling them to perform various NLP tasks effectively. Some notable transformer-based models include:
\begin{itemize}
    \item \textbf{Encoder-only Models}: BERT \cite{devlin2018bert} is designed for tasks requiring a deep understanding of the input text, such as sentiment analysis and question answering.
    \item \textbf{Encoder-Decoder Models}: BART \cite{lewis2020bart} and T5 \cite{raffel2020exploring} are suitable for sequence-to-sequence tasks like translation and text summarization.
    \item \textbf{Decoder-only Models}: The GPT family \cite{radford2019language} excels at text generation tasks, including chatbots and creative writing (see architectures of LLaMA and GPT-3 in\ref{fig:architectures}).
\end{itemize}

Recent advancements include scaling up these models to improve performance and efficiency. Techniques like LoRA \cite{hu2021lora} and parameter-efficient fine-tuning \cite{lialin2023scaling} have been introduced to make these large models more accessible and adaptable to specific tasks.

The transformer architecture, with its self-attention mechanism and ability to process entire sequences in parallel, represents a significant leap forward in NLP. Its flexibility and scalability have led to numerous successful applications across different domains, including finance \cite{wu2023bloomberggpt}, language instruction \cite{chung2024scaling}, and more. As research continues to refine these models, their capabilities are expected to expand further, offering even more sophisticated solutions to complex language tasks.

\section{LLM Training Resources: GPU Memory Requirements}

\subsection{Scaling Model Training Across Multiple GPUs}

\begin{figure}[tbp]
\centering
\includegraphics[width=0.7\textwidth]{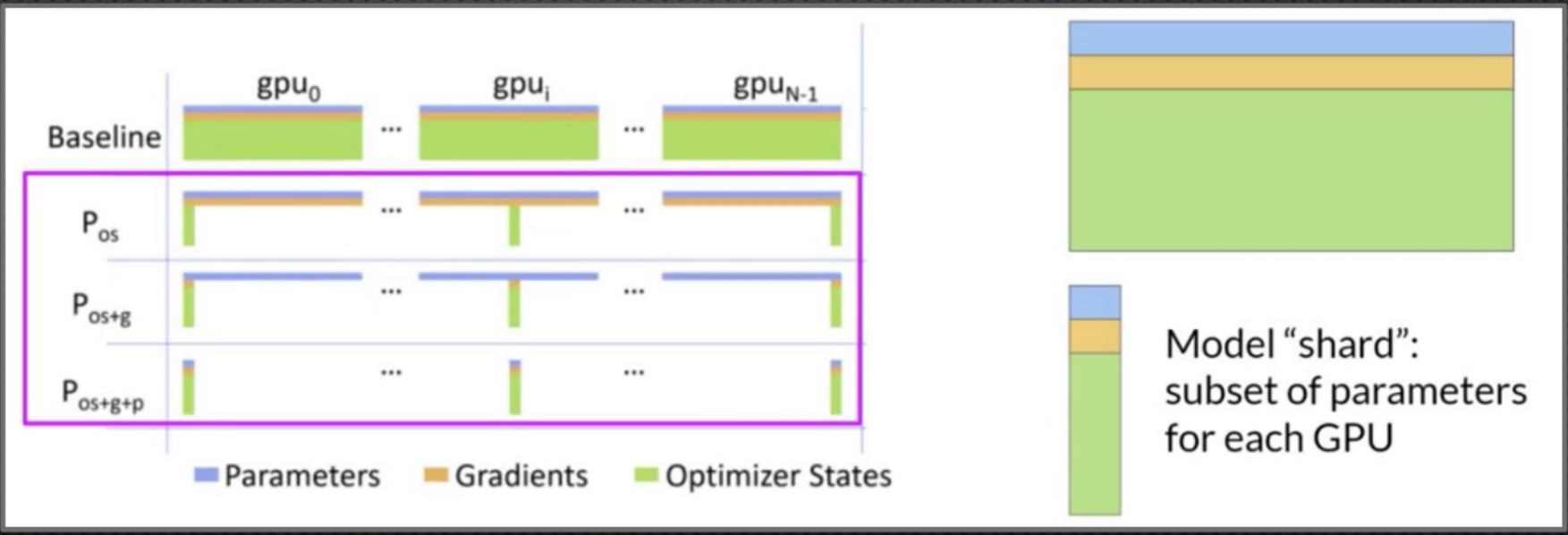}
\caption{Overview of the three stages of ZeRO optimization.}
\label{fig:FSDP}
\end{figure}

Scaling model training across multiple GPUs is essential for handling large models that cannot fit into a single GPU's memory. Even for models that can fit on one GPU, utilizing multiple GPUs can significantly accelerate training times. Efficiently distributing computation across GPUs is critical for optimizing performance. Initially, when a model fits onto a single GPU, the primary strategy for scaling involves distributing large datasets across multiple GPUs, enabling parallel processing of data batches. PyTorch's Distributed Data Parallel (DDP) is a popular method for this purpose. DDP replicates the entire model on each GPU, processes data batches in parallel, and synchronizes the results to ensure model consistency across GPUs. This method significantly speeds up training but requires that all model weights, parameters, gradients, and optimizer states fit within a single GPU's memory. When models exceed this memory capacity, model sharding techniques, such as Fully Sharded Data Parallel (FSDP), become necessary.
FSDP, inspired by Microsoft's 2019 ZeRO (Zero Redundancy Optimizer) paper \cite{rajbhandari2019zero}, addresses memory optimization by distributing model states across GPUs without redundancy. ZeRO introduces three stages of optimization (see Fig. \ref{fig:FSDP}):
\begin{itemize}
\item \textbf{ZeRO Stage 1:} This stage shards only the optimizer states across GPUs, reducing the memory footprint by up to four times. Optimizer states typically consume twice the memory of the model weights, making this a significant optimization.
\item \textbf{ZeRO Stage 2:} In addition to sharding optimizer states, this stage also shards gradients across GPUs. Combined with Stage 1, this can reduce the memory footprint by up to eight times, further optimizing memory usage and enabling training of larger models.
\item \textbf{ZeRO Stage 3:} This stage shards all model components, including model parameters, gradients, and optimizer states. Applied together with Stages 1 and 2, it offers a linear reduction in memory usage proportional to the number of GPUs, potentially reducing memory usage by a factor of 64 when sharding across 64 GPUs.
\end{itemize}
FSDP \cite{zhao2023pytorch} leverages these stages to distribute model parameters, gradients, and optimizer states across GPUs, avoiding the replication required by DDP. Unlike DDP, which maintains a complete copy of the model on each GPU, FSDP requires GPUs to share necessary data before forward and backward passes. This process involves temporarily reconstructing sharded data into unsharded form for the operation duration and then reverting to shards post-operation. This strategy reduces overall GPU memory usage but increases communication overhead, necessitating a trade-off between performance and memory. For smaller models, FSDP and DDP perform comparably. However, for larger models, DDP often encounters out-of-memory errors, whereas FSDP manages these models effectively, especially with 16-bit precision. Performance slightly decreases with an increasing number of GPUs due to higher communication volumes.
In summary, FSDP facilitates efficient scaling for both small and large models, enabling seamless training across multiple GPUs (see Fig. \ref{fig:FSDP}). Understanding how to distribute data, model parameters, and training computations is crucial for optimizing the training of large language models (LLMs). Future work will explore further performance enhancements, particularly with smaller models, as discussed in subsequent sections.

\subsection{The Era of 1-bit LLMs: Efficient and High-Performance Model Training}

The recent development of 1-bit Large Language Models (LLMs) \cite{ma2024era} marks a significant advancement in the efficiency and performance of model training and inference. Traditional LLMs rely on 16-bit floating point (FP16 or BF16) precision for computations, leading to high memory and energy demands. In contrast, the introduction of BitNet b1.58, a 1-bit LLM variant, represents a paradigm shift by quantizing model weights to ternary values {-1, 0, 1}, thereby significantly reducing computational and memory overhead while maintaining model performance. BitNet b1.58 offers several notable benefits:
\begin{itemize}
\item \textbf{Reduced Memory Footprint:} By representing weights with only 1.58 bits, BitNet b1.58 dramatically reduces the memory requirements compared to FP16 models. For example, a 3B parameter model using BitNet b1.58 consumes 3.55 times less GPU memory than its FP16 counterpart \cite{WMD+23}.
\item \textbf{Lower Latency and Higher Throughput:} The 1-bit model architecture significantly decreases inference latency and increases throughput. BitNet b1.58 achieves up to 2.71 times faster inference and 8.9 times higher throughput than FP16 models of the same size \cite{ma2024era}.
\item \textbf{Energy Efficiency:} The energy consumption for arithmetic operations is drastically reduced in BitNet b1.58 due to the use of integer addition instead of floating-point operations. This leads to substantial energy savings, with BitNet b1.58 being 71.4 times more energy-efficient in matrix multiplication operations \cite{ma2024era}.
\item \textbf{Comparable Performance:} Despite the lower precision, BitNet b1.58 matches or exceeds the performance of FP16 models in terms of perplexity and end-task accuracy, particularly for models starting from a 3B size \cite{ma2024era}.
\end{itemize}
BitNet b1.58 represents a significant advancement in the efficient training and deployment of LLMs. By reducing memory footprint, latency, and energy consumption while maintaining high performance, 1-bit LLMs like BitNet b1.58 provide a promising direction for the future of scalable and sustainable AI. This new line of research not only optimizes the current generation of LLMs but also sets the stage for designing specialized hardware optimized for 1-bit computations, further enhancing the feasibility of deploying large models in resource-constrained environments.

\section{Fine-Tuning Strategies}

For fine-tuning models, we often need multi-task settings and diverse datasets to ensure robust performance across various tasks. Having references to the available datasets is crucial for researchers and practitioners in selecting appropriate data for their specific fine-tuning needs. Table \ref{tab:datasets} presents an overview of datasets and benchmarks commonly used in natural language processing tasks, including their type, citation, and a brief description. The table provides a comprehensive list of these datasets and benchmarks, serving as a valuable resource for understanding the scope and capabilities of different data sources in the field of natural language processing.

\begin{table}[t]
\centering
\resizebox{\textwidth}{!}{ 
\begin{tabular}{|l|c|l|l|}
\hline
\textbf{Name} & \textbf{Type} & \textbf{Citation} & \textbf{Description} \\
\hline
GLUE & Benchmark & \cite{wang2018glue} & A multi-task benchmark for natural language understanding \\
SuperGLUE & Benchmark & \cite{wang2019superglue} & An improved version of GLUE with more challenging tasks \\
MMLU & Benchmark & \cite{hendrycks2020measuring} & Massive multitask language understanding benchmark \\
BIG-Bench & Benchmark & \cite{suzgun2022challenging} & A benchmark for testing LLMs on diverse tasks \\
HELM & Benchmark & \cite{liang2022holistic} & Holistic evaluation of language models \\
Gemini & Benchmark & \cite{saab2024gemini} & Evaluates capabilities of LLMs in medicine \\
CoLA & Dataset & \cite{warstadt2019neural} & Corpus of Linguistic Acceptability \\
SST2 & Dataset & \cite{socher2013recursive} & Stanford Sentiment Treebank, binary sentiment classification \\
MRPC & Dataset & \cite{dolan2005automatically} & Microsoft Research Paraphrase Corpus \\
STS & Dataset & \cite{cer2017semeval} & Semantic Textual Similarity benchmark \\
QQP & Dataset & \cite{iyer2017first} & Quora Question Pairs \\
MNLI & Dataset & \cite{williams2018broad} & Multi-Genre Natural Language Inference \\
QNLI & Dataset & \cite{wang2018glue} & Question Natural Language Inference \\
RTE & Dataset & \cite{bentivogli2009fifth} & Recognizing Textual Entailment \\
WNLI & Dataset & \cite{levesque2012winograd} & Winograd NLI \\
CoT & Dataset & \cite{wei2022chain} & Chain-of-Thought Reasoning \\
Muffin & Dataset & \cite{singh2022muffin} & Dataset for fine-tuning LLMs on diverse tasks \\
Natural Instructions v2 & Dataset & \cite{mishra2022cross} & Collection of natural language instructions \\
MGSM & Dataset & \cite{cobbe2021measuring} & Measuring generalization and systematicity in models \\
BBH & Dataset & \cite{srivastava2022beyond} & Beyond Big-Bench: Dataset for testing generalization \\
\hline
\end{tabular}}
\caption{Overview of datasets and benchmarks for natural language processing tasks.}\label{tab:datasets}
\end{table}

\subsection{Improving Performance of Large Language Models through Fine-Tuning}

In this section, we explore methodologies for enhancing the performance of Large Language Models (LLMs) tailored to specific applications. Fine-tuning is a pivotal technique in this enhancement process, involving the adjustment of a pre-trained model using a dataset of labeled examples to improve task-specific performance. Fine-tuning diverges from pre-training, which leverages extensive amounts of unstructured textual data for self-supervised learning. In contrast, fine-tuning employs supervised learning, utilizing prompt-completion pairs to update the model's weights. A particularly effective variant of this method is instruction fine-tuning. Instruction fine-tuning utilizes examples that explicitly demonstrate the desired model responses to specific instructions. Prompts such as "identify the sentiment of this tweet" or "translate this sentence into Spanish" guide the model to generate appropriate completions. The training dataset comprises numerous such pairs, enabling the model to learn to generate responses according to given instructions.

Fine-tuning can be resource-intensive, demanding substantial memory and compute capacity to process gradients and optimizers. However, advancements in memory optimization and parallel computing strategies can alleviate these demands \cite{shoeybi2019megatron}. To fine-tune an LLM, an appropriately curated training dataset is essential. Publicly available datasets, though often not in an instruction format, can be transformed using prompt template libraries. These libraries provide templates for various tasks and datasets, such as Amazon product reviews, to create instruction-based datasets. For example, templates can generate prompts like "determine the relevance of this news article" or "provide a brief summary of this research abstract," integrating instructions with dataset examples. Upon preparing the instruction dataset, it is partitioned into training, validation, and test splits. During fine-tuning, prompts from the training set are used to generate completions, which are compared against the expected responses. The model outputs a probability distribution across tokens, and the cross-entropy loss function calculates the discrepancy between the generated and expected distributions. This loss is then used to update the model's weights through backpropagation \cite{goodfellow2016deep}. The fine-tuning process involves multiple batches and epochs, continuously updating the model's weights to enhance performance. Evaluation using holdout validation and test datasets yields metrics such as validation accuracy and test accuracy, providing a measure of the model's performance. The result of fine-tuning is an improved version of the base model, referred to as an instruct model, optimized for specific tasks. Instruction fine-tuning is the predominant method for fine-tuning LLMs today, typically implied when discussing fine-tuning in this context.

Large Language Models (LLMs) are renowned for their versatility in performing various language tasks within a single model. However, specific applications often necessitate excelling in a single task. Fine-tuning a pre-trained model on a particular task, such as summarization, can significantly enhance performance, achievable with relatively small datasets, often just 500-1,000 examples, compared to the billions of texts used during pre-training \cite{radford2019language, brown2020language}. Despite its advantages, fine-tuning on a single task may induce catastrophic forgetting, a phenomenon where the model's performance on previously learned tasks deteriorates. This occurs as fine-tuning optimizes the model's weights for the new task, potentially impairing its ability to perform other tasks. For example, while fine-tuning might improve a model's sentiment analysis capabilities, it could diminish its effectiveness in named entity recognition \cite{mcclelland1995there}. To mitigate catastrophic forgetting, one must assess its impact on their specific use case. If the primary requirement is reliable performance on the fine-tuned task, the loss of multitask capabilities may be acceptable. However, if maintaining generalized capabilities is essential, multitask fine-tuning offers an alternative. This approach involves fine-tuning on multiple tasks simultaneously, requiring more data (50,000-100,000 examples) and computational resources \cite{caruana1997multitask}. Table \ref{tab:datasets} presents an overview of datasets and benchmarks commonly used in natural language processing tasks, including their type, citation, and a brief description.
 Another strategy is parameter-efficient fine-tuning (PEFT), which preserves the original LLM weights by training only a small number of task-specific adapter layers. PEFT is more robust to catastrophic forgetting as it retains most of the pre-trained weights unchanged. This technique is an active research area and shows promising results in maintaining multitask performance while fine-tuning \cite{houlsby2019parameter}.

In summary, while fine-tuning can enhance performance for specific tasks, it is crucial to consider the trade-offs and potential strategies to avoid catastrophic forgetting based on the application's requirements.

\subsection{Multitask Fine-Tuning and Instruction-Tuned Models}

Multitask fine-tuning extends single-task fine-tuning by employing a dataset containing inputs and outputs for multiple tasks, such as summarization, review rating, code translation, and entity recognition. This approach allows the model to improve performance across all tasks simultaneously while mitigating catastrophic forgetting. During training, the model's weights are updated based on the cumulative losses over numerous epochs, resulting in a model proficient in multiple tasks. However, multitask fine-tuning is data-intensive, often requiring 50,000 to 100,000 examples. The FLAN (Fine-tuned Language Net) models exemplify the effectiveness of multitask instruction fine-tuning. These models, such as FLAN-T5, are fine-tuned on a large and diverse set of datasets, achieving robust performance across various tasks. For instance, FLAN-T5 has been fine-tuned on 473 datasets spanning 146 task categories. The process leverages datasets like SAMSum, which includes 16,000 messenger-like conversations with summaries crafted by linguists.

In training language models for summarization tasks, prompt templates are applied to the SAMSum dialogues to instruct the model to summarize conversations. These templates, varying in phrasing but consistent in purpose (e.g., "briefly summarize the dialogue" or "provide a summary of this dialogue"), help the model generalize better. Despite the general capabilities of FLAN-T5, additional fine-tuning on domain-specific datasets can enhance its performance for particular use cases. For example, the DialogSum dataset, containing over 13,000 support chat dialogues and summaries, can improve the model's ability to summarize customer service interactions. This involves fine-tuning the model with dialogues relevant to the target domain, leading to more accurate and pertinent summarization. An example of this process is seen with a support chat from the DialogSum dataset, where initial summaries by FLAN-T5 may lack critical details or contain inaccuracies. Post fine-tuning on the DialogSum dataset, the model's summaries align more closely with human-generated ones, capturing all essential details without introducing inaccuracies. Fine-tuning with specific datasets, including internal company data, ensures the model learns the nuances of the required tasks, yielding more accurate and useful outputs for applications like customer support. Evaluating the model's output quality is crucial, and various metrics and benchmarks are employed to assess performance improvements post-fine-tuning. For a list of benchmarks and dataset see Table \ref{tab:datasets}.

\subsection{Parameter-Efficient Fine-Tuning (PEFT)}

Training Large Language Models (LLMs) is a resource-intensive task requiring substantial computational and memory resources. Full fine-tuning updates all model parameters, necessitating storage not only for the model but also for optimizer states, gradients, and activations. The memory requirements can quickly surpass the capacity of consumer hardware, as models and training components can be many times larger than the model itself \cite{brown2020language}.

Parameter-Efficient Fine-Tuning (PEFT) offers a solution by updating only a subset of parameters, significantly reducing the memory footprint. PEFT methods can be categorized into selective, reparameterization, and additive techniques.

Selective methods fine-tune only specific components or layers of the model. Reparameterization techniques, such as Low-Rank Adaptation (LoRA), introduce low-rank decompositions of weight matrices, reducing the number of parameters that need training \cite{hu2021lora}. Additive methods, including adapters and soft prompts, introduce new trainable components while keeping the original model weights frozen.

\subsubsection{Low-Rank Adaptation (LoRA)}

\begin{figure}[tbp]
\centering
\includegraphics[width=0.7\textwidth]{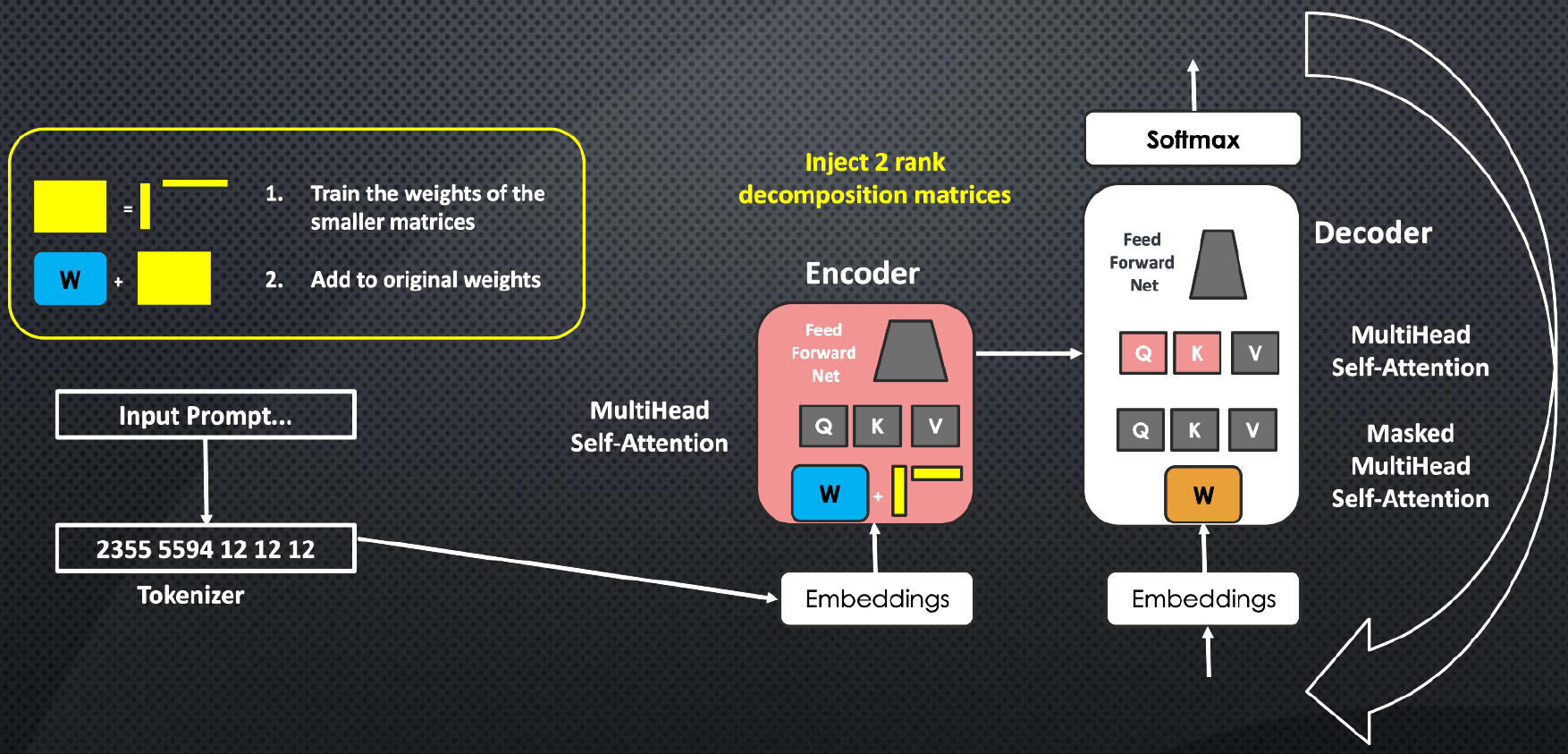}
\caption{LoRA overview, showing the injection of low-rank matrices.}
\label{fig:lora}
\end{figure}

LoRA is a reparameterization technique that freezes the original LLM weights and injects low-rank decomposition matrices alongside them. This approach significantly reduces the number of trainable parameters by creating small matrices whose product approximates the original weights (see Fig. \ref{fig:lora}). LoRA employs a decomposition where a large weight matrix \( W \in \mathbb{R}^{d \times k} \) is approximated by the product of two smaller matrices \( A \in \mathbb{R}^{d \times r} \) and \( B \in \mathbb{R}^{r \times k} \), where \( r \ll \min(d, k) \). This decomposition can be expressed as:
\[
W \approx A \cdot B
\]
During fine-tuning, only the matrices \( A \) and \( B \) are updated, while \( W \) remains fixed. This reduces the number of parameters to \( r \times (d + k) \), which is much smaller than \( d \times k \). For example, if \( d = 512 \), \( k = 64 \), and \( r = 8 \), the reduction in parameters is substantial, from 32,768 to 4,608 \cite{hu2021lora}.

LoRA's primary advantage is its ability to drastically reduce the number of trainable parameters. Despite this reduction, models fine-tuned with LoRA maintain high performance. The technique leverages the fact that the essential information in the weight matrices can often be captured with a lower-rank approximation, thus preserving the model's ability to perform well on downstream tasks. For instance, fine-tuning the FLAN-T5 model for dialogue summarization using LoRA demonstrated significant performance improvements with minimal computational resources, achieving comparable ROUGE scores to full fine-tuning \cite{hu2021lora}.

Selecting the appropriate rank for LoRA matrices is crucial for balancing parameter reduction and model performance. Research suggests that ranks in the range of 4-32 offer a good trade-off, with minimal performance loss beyond a rank of 16 \cite{hu2021lora}. This optimization remains an active research area, with ongoing studies refining best practices.

\subsubsection{The Power of Scale for Parameter-Efficient Prompt Tuning}

\begin{figure}[tbp]
\centering
\includegraphics[width=0.7\textwidth]{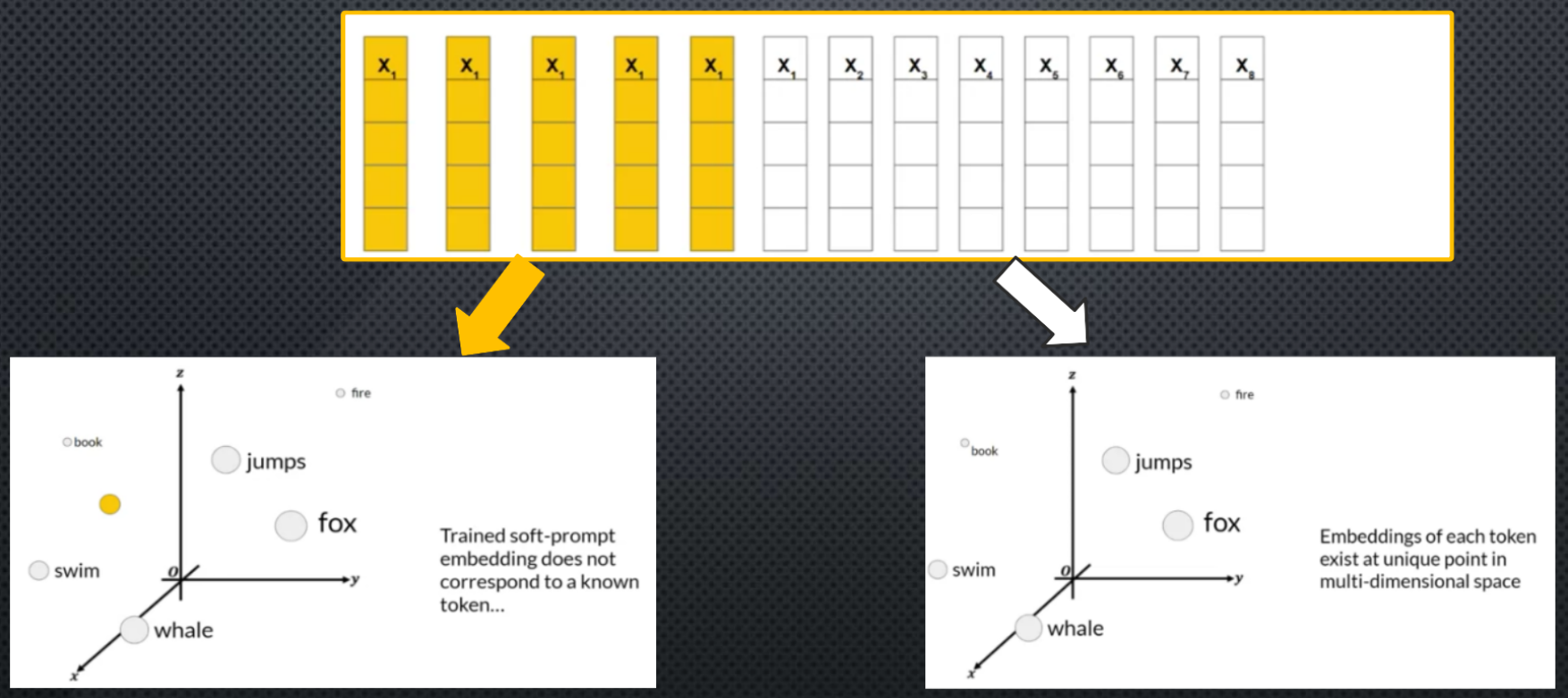}
\caption{Soft Prompts overview, showing the addition of trainable tokens.}
\label{fig:softprompts}
\end{figure}

Prompt tuning involves adding trainable tokens (soft prompts) to the input text. Unlike prompt engineering, which manually adjusts the language of prompts for better completions, prompt tuning allows supervised learning to optimize these additional tokens. Soft prompts, which are virtual tokens in the continuous embedding space, are prepended to the input text's embedding vectors. During training, these soft prompts are adjusted while keeping the model weights fixed. This approach is highly parameter efficient, as it only involves a small number of additional parameters compared to the millions or billions in full fine-tuning (see Fig. \ref{fig:softprompts}) \cite{lester2021prompt}.

Soft prompts are represented as additional embedding vectors \( P \in \mathbb{R}^{p \times e} \), where \( p \) is the length of the prompt and \( e \) is the embedding dimension. These prompts are concatenated to the embedded input matrix \( X_e \in \mathbb{R}^{n \times e} \), forming a new input matrix \( [P; X_e] \in \mathbb{R}^{(p+n) \times e} \). The model then processes this combined input as usual. Only the parameters of \( P \) are updated during training:
\[
\text{New input} = [P; X_e]
\]

Training involves backpropagation to adjust the parameters of the soft prompts. This method leverages the large pre-trained model and adapts it to new tasks with minimal additional parameters. For instance, tuning a 5-token prompt for a model with an embedding dimension of 2048 only adds 10,240 parameters \cite{lester2021prompt}.

Prompt tuning's effectiveness varies with model size. Lester et al. \cite{lester2021prompt} demonstrated that while prompt tuning underperforms full fine-tuning for smaller models, it becomes competitive as model size increases, particularly for models with around 10 billion parameters. This efficiency and scalability make prompt tuning a viable alternative to full fine-tuning, especially when computational resources are constrained.

Despite their virtual nature, analysis shows that the nearest neighbor tokens to soft prompts form semantically meaningful clusters. This suggests that soft prompts learn representations akin to word meanings relevant to the given task. For example, prompts trained on a dataset like BoolQ often have nearest neighbors related to the task's domain, indicating their role in priming the model to interpret inputs within a specific context \cite{lester2021prompt}.

In summary, both LoRA and prompt tuning exemplify PEFT methods that balance performance and computational efficiency. LoRA, with its use of rank decomposition matrices, and prompt tuning, with its trainable tokens, reduce the compute and memory requirements for fine-tuning large language models. These techniques are instrumental in optimizing fine-tuning processes, allowing for significant cost and resource savings while maintaining or enhancing model performance across various tasks.

\subsubsection{Comparative Analysis and Further Research}

The comparative analysis of LoRA and prompt tuning methods highlights their complementary strengths in parameter-efficient fine-tuning. LoRA's decomposition approach is particularly effective for scenarios where significant parameter reduction is essential without compromising model performance. Its matrix factorization technique allows for a substantial reduction in trainable parameters, making it feasible to fine-tune large models on hardware with limited memory capacity. Additionally, LoRA's ability to maintain high performance despite the reduced parameter count makes it a valuable tool for deploying LLMs in resource-constrained environments.

On the other hand, prompt tuning leverages the power of soft prompts to condition pre-trained models for specific tasks with minimal additional parameters. This method is highly scalable and effective for large models, where the addition of a small number of trainable tokens can significantly influence the model's behavior. The simplicity of prompt tuning, combined with its ability to retain the pre-trained model's knowledge, makes it an attractive option for fine-tuning large models for a variety of tasks.

Future research in PEFT techniques should focus on optimizing the balance between parameter efficiency and model performance. One promising direction is to explore hybrid approaches that combine the strengths of LoRA and prompt tuning. Such methods could leverage low-rank matrix decomposition to reduce the parameter count while utilizing soft prompts to enhance task-specific performance. Additionally, investigating the applicability of PEFT methods across different model architectures and domains will provide insights into their versatility and generalizability.

Another important aspect of future research is the development of adaptive PEFT methods that can dynamically adjust the parameter efficiency based on the complexity and requirements of the task. By incorporating mechanisms for task-aware adaptation, these methods could further enhance the fine-tuning process and improve the overall performance of LLMs in diverse applications.

In conclusion, parameter-efficient fine-tuning techniques such as LoRA and prompt tuning represent significant advancements in the field of LLM optimization. Their ability to reduce the computational and memory requirements of fine-tuning processes makes them essential tools for deploying large models in practical settings. As research in this area progresses, the development of more efficient and effective PEFT methods will continue to drive the evolution of LLMs, enabling their widespread adoption and application across various domains.

\section{Comparison of Reinforcement Learning from Human Feedback (RLHF) and Reinforced Self-Training (ReST)}

Reinforcement Learning from Human Feedback (RLHF) and Reinforced Self-Training (ReST) are two advanced methodologies designed to enhance the alignment of large language models (LLMs) with human preferences. This section provides a comprehensive overview of RLHF, detailing the reward model mechanism, introduces the innovative ReST approach, and conducts a comparative analysis of the two techniques in terms of principles, efficiency, and application outcomes.

RLHF \cite{ziegler2019,glaese2022improving} is a fine-tuning method that leverages human feedback to iteratively train language models, ensuring their outputs align closely with human expectations. RLHF aligns models with human preferences, ensuring outputs are relevant and minimize harm, such as avoiding toxic language \cite{christiano2017}. The process involves human annotators evaluating the model's outputs, ranking them based on quality and alignment with desired criteria. A reward model is then trained using this human feedback, assigning scores to outputs based on their alignment with human preferences. Subsequently, the language model (policy) is fine-tuned using reinforcement learning algorithms such as Proximal Policy Optimization (PPO) \cite{schulman2017proximal} and Asynchronous Advantage Actor-Critic (A2C), guided by the reward model. The core principle of RLHF involves the reward model's iterative optimization of the language model's policy. By continuously sampling outputs, scoring them with the reward model, and updating the policy, RLHF aims to enhance the model's performance on tasks as evaluated by human feedback \cite{stiennon2020learning}. However, RLHF faces several challenges, including substantial computational costs due to the need for ongoing interaction between the model and the reward system, and the risk of reward hacking, where models might exploit the reward model, producing outputs that maximize reward scores without genuinely improving quality.

Reinforced Self-Training (ReST) is a novel approach proposed to address some limitations of traditional RLHF by integrating reinforcement learning with self-training. The primary innovation of ReST lies in its two-loop structure: the Grow step and the Improve step. In the Grow step, the model generates a large dataset of output predictions from its current policy, similar to data augmentation in self-training. The Improve step involves filtering and ranking the generated dataset using a scoring function, often a reward model trained on human feedback, and then fine-tuning the model on this filtered dataset using offline reinforcement learning algorithms. This approach has several advantages over traditional RLHF. By separating the data generation and policy improvement phases, ReST reduces the computational burden. The dataset generated in the Grow step can be reused in multiple Improve steps, making the process more efficient. The decoupled nature of ReST allows for careful inspection and filtering of the generated data, mitigating issues like reward hacking. Moreover, ReST tends to be more stable with fewer hyperparameters to tune compared to online RL methods.

Comparative analysis reveals distinct advantages and limitations for each approach. In terms of efficiency and computational cost, RLHF requires continual interaction with the reward model, leading to high computational costs due to the need for constant sampling and scoring of new data. Conversely, ReST generates a dataset once and reuses it, significantly reducing the computational load. This makes ReST more suitable for large-scale applications where computational resources are a constraint. Regarding the quality of outputs, RLHF's direct influence from the reward model and feedback loop can lead to high-quality improvements, but also opens the door to reward hacking. ReST benefits from its offline RL approach, which allows for better control over data quality. The use of a fixed dataset for improvement steps helps maintain consistency and reduce the risk of reward hacking.

In terms of application and generalizability, RLHF is broadly applicable across various tasks, but its effectiveness can be limited by the scalability of the reward model and the computational costs involved. ReST is particularly effective in settings where high-quality reward models are available and can be used to generate robust datasets for offline training. Its general approach can be applied to any generative learning setting, making it versatile and scalable. Experimental results indicate that RLHF has shown significant improvements in tasks like summarization and translation, but these improvements often come at the cost of increased computational resources. ReST has demonstrated substantial improvements in machine translation benchmarks with less computational effort. For instance, experiments with the IWSLT 2014 and WMT 2020 benchmarks showed that ReST could improve translation quality significantly as measured by automated metrics and human evaluation.

The theoretical foundations and practical implications of each method further underscore their unique advantages. RLHF is grounded in online reinforcement learning principles, with human feedback guiding the learning process. This real-time interaction makes it powerful but also prone to instability and exploitation. ReST builds on the principles of offline RL and self-training, combining them into a more structured and efficient methodology. This approach enhances stability and ensures that model improvements are grounded in robust, human-evaluated data.

In conclusion, both RLHF and ReST represent significant strides in the field of language model alignment, each with unique strengths and weaknesses. RLHF's direct use of human feedback in an online learning framework makes it highly responsive to human preferences, albeit at a higher computational cost and risk of reward hacking. ReST, by decoupling data generation and policy improvement, offers a more efficient and stable alternative, particularly suitable for large-scale applications. Future research could explore hybrid approaches that leverage the strengths of both methods, potentially leading to even more robust and efficient language model training paradigms.

\subsection{Proximal Policy Optimization for Large Language Models}

Proximal Policy Optimization (PPO) is a reinforcement learning algorithm designed to optimize policies by iteratively updating the model within a bounded region, ensuring stability and improved performance. In the context of Large Language Models (LLMs), PPO aligns the model more closely with human preferences over successive iterations \cite{schulman2017proximal}. The PPO algorithm operates in two phases. Phase I involves the LLM conducting a series of tasks to complete given prompts, generating responses that are evaluated against a reward model capturing human preferences. Rewards are quantified based on criteria such as helpfulness, harmlessness, and honesty. The expected reward of a completion, estimated by a value function, is a critical component in the PPO objective. In Phase II, the weights of the LLM are updated based on the prompt completions' losses and rewards from Phase I. These updates are constrained within a trust region to ensure the new policy remains close to the initial one, thus maintaining stability. The main objective in this phase is to optimize the policy loss, which balances maximizing rewards and maintaining policy proximity. The PPO objective function can be expressed as:
\[
L^{CLIP}(\theta) = \mathbb{E}_t \left[ \min \left( r_t(\theta) \hat{A}_t, \text{clip}(r_t(\theta), 1 - \epsilon, 1 + \epsilon) \hat{A}_t \right) \right],
\]
where \( r_t(\theta) \) is the probability ratio between the new and old policies, and \( \hat{A}_t \) is the advantage estimate for a given action. The clipping function ensures updates remain within a small, stable region, referred to as the trust region, mitigating the risk of substantial deviations that could destabilize learning \cite{schulman2017proximal}.

Entropy loss is also integrated into the PPO objective to preserve model creativity, balancing the policy's alignment with human preferences and its generative diversity. This entropy component operates similarly to the temperature setting in LLM inference, but it influences creativity during training. The complete PPO objective combines the policy loss, value function loss, and entropy loss, weighted by hyperparameters \( C_1 \) and \( C_2 \):

\[
L(\theta) = \mathbb{E}_t \left[ L^{CLIP}(\theta) - C_1 L^{VF}(\theta) + C_2 S[\pi](s_t) \right],
\]

where \( L^{VF}(\theta) \) is the value function loss and \( S[\pi](s_t) \) represents the entropy term. Through backpropagation, the model weights are updated iteratively, refining the LLM to align more closely with human preferences. While PPO is widely used for Reinforcement Learning with Human Feedback (RLHF), other techniques like Q-learning are also explored for fine-tuning LLMs. However, PPO's balance of complexity and performance makes it a prevalent choice. Research in this area is ongoing, with new methods like Direct Preference Optimization offering promising alternatives \cite{christiano2017deep, ouyang2022training}.

\section{Conclusion}
This tutorial paper has explored advancements and challenges in the development of Large Language Models (LLMs) like ChatGPT and Gemini. These models face limitations such as outdated knowledge, difficulty in complex computations, and generating incorrect information. To address these, several innovative techniques and frameworks have been introduced. Retrieval Augmented Generation (RAG) connects LLMs to current external databases, enhancing their accuracy and relevance. Program-Aided Language Models (PAL) use external code interpreters for precise computations, broadening LLM capabilities. LangChain, an open-source framework, simplifies the integration of LLMs with external data sources, enabling the development of domain-specific applications efficiently. It supports diverse applications such as chatbots and content generation without needing retraining or fine-tuning.
Fine-tuning strategies like instruction fine-tuning, multitask fine-tuning, and parameter-efficient methods such as Low-Rank Adaptation (LoRA) and prompt tuning are discussed to mitigate catastrophic forgetting and improve performance.
Reinforcement Learning from Human Feedback (RLHF) and Reinforced Self-Training (ReST) align LLMs with human preferences. RLHF uses human evaluations for iterative fine-tuning, while ReST combines reinforcement learning with self-training for efficiency and reduced computational costs. These methods refine LLM outputs to better meet user expectations. The paper also reviews transformer architectures that have revolutionized NLP, highlighting recent advancements for performance and efficiency improvements. Techniques for scaling model training beyond a single GPU, such as PyTorch’s Distributed Data Parallel (DDP) and Fully Sharded Data Parallel (FSDP), along with the ZeRO stages for memory optimization, are discussed.
Practical applications, such as customer service bots, demonstrate the benefits of integrating LLMs with real-time data and advanced reasoning strategies. These integrations enable LLMs to provide accurate, contextually relevant responses, enhancing user interactions. In summary, this tutorial presents a comprehensive examination of state-of-the-art advancements and challenges in LLM development. It introduces innovative techniques and frameworks to enhance LLM performance, reliability, and applicability. The overview of transformer architectures, fine-tuning strategies, and integration frameworks offers valuable insights for future research and practical applications in NLP.
The source code can be accessed by contacting the author via email for a request.

\textit{If you use the concepts presented in this paper for your research or any other purposes, please cite our work}.\\

\bibliographystyle{plain}
\bibliography{references}

\end{document}